\documentclass[pdflatex,sn-mathphys-num]{sn-jnl}
\usepackage{booktabs}
\usepackage{longtable}
\usepackage{pbox}
\usepackage{graphicx}%
\usepackage{multirow}%
\usepackage{amsmath,amssymb,amsfonts}%
\usepackage{amsthm}%
\usepackage{mathrsfs}%
\usepackage[title]{appendix}%
\usepackage{xcolor}%
\usepackage{textcomp}%
\usepackage{manyfoot}%
\usepackage{booktabs}%
\usepackage{algorithm}%
\usepackage{algorithmicx}%
\usepackage{algpseudocode}%
\usepackage{listings}%

\theoremstyle{l}%

\theoremstyle{thmstyleone}%

\theoremstyle{thmstyletwo}%

\theoremstyle{thmstylethree}%

\begin{document}

\title[Article Title]{Memory Merge DQN: Sensitivity Weighted Target Updates for Stable Value Learning}

\author*[1]{\fnm{Adrian} \sur{Ly}}\email{lyadr@deakin.edu.au}

\author[1]{\fnm{Richard} \sur{Dazeley}}\email{richard.dazeley@deakin.edu.au}
\equalcont{These authors contributed equally to this work.}

\author[2]{\fnm{Peter} \sur{Vamplew}}\email{p.vamplew@federation.edu.au}
\equalcont{These authors contributed equally to this work.}

\author[1]{\fnm{Sunil} \sur{Aryal}}\email{sunil.aryal@deakin.edu.au}
\equalcont{These authors contributed equally to this work.}

\author[3]{\fnm{Francisco} \sur{Cruz}}\email{f.cruz@unsw.edu.au}
\equalcont{These authors contributed equally to this work.}

\affil*[1]{\orgdiv{Information Technology}, \orgname{Deakin University}, \orgaddress{\street{Warun Ponds}, \city{Geelong}, \postcode{3220}, \state{Victoria}, \country{Australia}}}

\affil[2]{\orgdiv{Institute of Innovation, Science and Sustainability}, \orgname{Federation University}, \orgaddress{\street{Mt Helen}, \city{Ballarat}, \postcode{3350}, \state{Victoria}, \country{Australia}}}

\affil[3]{\orgdiv{Information Technology}, \orgname{UNSW}, \orgaddress{\street{Kensington}, \city{Sydney}, \postcode{2033}, \state{New South Wales}, \country{Australia}}}

\abstract{Deep Q-networks use target networks to stabilise bootstrapped value learning, but the standard hard copy update also introduces a tradeoff. Holding the target network fixed, improves short term stability, yet each hard update abruptly replaces the target parameters with the newest online network and discards recent parameter history. This can produce sudden changes in the bootstrap target and may remove value function structure that remains useful later in training. This paper introduces Memory Merge DQN, a target network update mechanism that maintains a short memory of recent historical online network copies and constructs the target network by merging network parameters based on the Q-value sensitivity rather than copying only the newest online network. Memory Merge gives greater influence to parameters that remain locally important for current Q-value behaviour, while using a recency prior to keep the merged target close to the latest online parameters. The method is inspired by Fisher Weight Model Merging, but uses Q-value sensitivity rather than Fisher information as the weighting signal. This paper evaluates Memory Merge DQN on Atari environments against DQN, Averaged DQN, DQN with layer normalisation, and PQN (with gradient clipping). The results show that Memory Merge DQN is highly competitive and it achieves the largest number of first place final performance results among the evaluated methods, beats DQN, Averaged DQN, and PQN (with gradient clipping), and produces substantial gains in several games where preserving useful value-function parameters appears beneficial. These findings suggest that selectively merging recent parameter weights and history can improve the stability and final performance of DQN agents, and that target network design is an important mechanism for preserving useful value function structure during long horizon value learning.}


\maketitle

\section{Introduction}

Value-based deep reinforcement learning methods such as Deep Q-networks (DQN) have shown that Q-learning can be scaled to high dimensional visual control tasks when combined with function approximation, experience replay, and a target network \citep{mnih2015human, mnih2013playing}. The target network is a central stabilising mechanism in DQN \cite{fan2020theoretical, hernandez2019understanding}. Rather than using the online network to both choose the current prediction and define the bootstrap target at every update, DQN maintains a separate target network whose parameters are held fixed for a number of training updates before being replaced by a copy of the online network. This temporary separation slows the movement of the bootstrapped target and helps reduce one source of instability in temporal difference learning \citep{mnih2015human, van2018deep}. During each target update interval, the target network becomes increasingly stale relative to the online network. When the hard update occurs, the target network is replaced abruptly, and the standard target network discards intermediate parameter history and introduces a sudden change in the learning target. Prior work has studied the implications of the hard update and have made relationships between this and several related phenomena, including capacity loss, dormant neurons, and primacy bias, where agents become overly shaped by early experience and respond less effectively later \citep{lyle2022understanding, sokar2023dormant, nikishin2022primacy}. 

These plasticity related failures are distinct from value overestimation, but the two problems can interact in value-based learning \cite{nikishin2022primacy,nikishin2018improving}. DQN bootstrap targets can produce overestimated action values under some conditions \citep{van2018deep}, and inaccurate value estimates copied into the target network can continue to influence subsequent temporal difference targets. From this perspective, the central problem considered in this paper is whether target networks can preserve useful value function structure without retaining outdated or misleading parameter information. Existing approaches provide only partial answers to this problem. Periodic reset methods can restore plasticity \citep{nikishin2022primacy, sokar2023dormant} by discarding part of the learned network, but this is a strong intervention that may also remove useful information. Soft target updates, such as Polyak averaging, move the target parameters gradually toward the online parameters and are widely used in actor-critic methods \citep{lillicrap2015continuous, fujimoto2018addressing, haarnoja2018soft}. These updates reduce abrupt target changes, but they apply a scalar smoothing rule uniformly across parameters. Averaged DQN instead averages the output Q-values of previously learned networks, reducing approximation error variance in the bootstrap target \citep{anschel2017averaged}. This preserves historical value estimates at the prediction level, but it does not construct a single target network by selectively retaining parameters according to their current functional importance. 

This paper proposes Memory Merge DQN, a target network update rule that gives the target network access to a short memory of recent online network states. Instead of replacing the target network with only the newest online parameters, Memory Merge stores the most recent $K$ online network copies and constructs the next target network through a Q-value sensitivity weighted parameter merge:

\[ \theta^- \leftarrow \operatorname{Merge}_{\mathrm{QS}} \left( \theta_{t_1}, \ldots, \theta_{t_K} \right). \] 

The merge rule is inspired by Fisher Weighted Model Merging \citep{matena2022merging}, but it does not estimate a Fisher information matrix. Instead, it uses squared gradients of selected Q-values with respect to each parameter as a sensitivity measure. Parameters that have greater influence on the current Q-value behaviour receive more weight in the merged target, while an additional recency prior keeps the target close to the newest online network. The aim is to preserve useful recent value function structure while avoiding a purely uniform average over all stored copies. The main hypothesis is that target network memory can improve value-based learning when the retained parameter history remains compatible with the current value function. 

This paper makes four contributions: 

\begin{itemize} 

\item Introduce Memory Merge DQN, a target network update mechanism that replaces hard target copying with a Q-value sensitivity weighted merge over recent online network copies. 

\item Adapt the idea of importance weighted parameter merging to bootstrapped value learning, using Q-value sensitivity as a measure of importance during the target update. 

\item Evaluate Memory Merge DQN on Atari environments against DQN, Averaged DQN, DQN with layer normalisation, and PQN (with gradient clipping), using per game rankings, pairwise comparisons, and large improvement counts rather than raw cross game score averages. 

\item Analyse when target network memory is beneficial by reporting a memory size ablation and by comparing predicted Q-value scale with episodic return as a diagnostic of value calibration. 

\end{itemize} 

Empirically, Memory Merge DQN is highly competitive, it achieves the largest number of first place final performance results in the evaluated games and outperforms vanilla DQN and is highly competitive with Parallelised Q-Networks (PQN). This pattern supports the view that selective target network design can produce substantial gains when recent parameter history remains useful.

\section{Related works}

DQN demonstrated that value-based reinforcement learning could be scaled to high dimensional pixel-based environments through the combination of Q-learning with function approximation, experience replay, and a target network \citep{mnih2015human}. The use of a target network in DQN addresses one of the central instabilities of bootstrapped updates, where the target that is used to train the online network relies on the same learned value function that is being updated. In standard DQN, this is mitigated by holding the target network parameters fixed for a number  of updates before copying the online network parameters into the target network. \citet{van2018deep} in their analysis of the deadly triad in DQN, examined how mechanisms like the target network can slow down divergence by the copying period. 

One of the most common methods of target updates is the Polyak averaging update, in which the target parameters are moved incrementally towards the online parameters at each update. This approach was adopted in Deep Deterministic Policy Gradients \citep{lillicrap2015continuous} and has since become standard in modern actor critic methods such as TD3 and SAC \citep{fujimoto2018addressing, haarnoja2018soft}. In contrast to hard copying, soft target updates reduce abrupt changes in bootstrap targets by imposing temporal smoothing  on the target parameters. However, the coefficient used to smooth the copying process is typically scalar and manually chosen, so all parameters are updated at the same rate regardless of their functional importance to the current value estimate.

The work most closely related to this paper's approach within the value-based domain of algorithms is Averaged DQN \citep{anschel2017averaged}. Averaged DQN maintains a series of previously learned Q-functions and averages their Q-value estimates when constructing targets. This reduces approximation error variance in the target values and leads to more stable training. This paper's method shares the motivation of using historical value function information to stabilise DQN, but differs in the object being averaged and in the weighting rule. Averaged DQN averages Q-value predictions from previous networks, whereas the method used in this paper constructs a single merged target network by averaging parameters of recent online network copies. Moreover, instead of using a uniform average over historical estimates, Memory Merge is parameter based and weighted by Q-value sensitivity on sample experiences from the replay buffer.

An alternative to target network based DQN is PQN \citep{gallici2024simplifying}. PQN revisits deep temporal difference learning and argues that stable Q-learning can be obtained without the standard DQN combination of a replay buffer and target network, provided that online data collection is sufficiently parallelised and the network is stabilised through layer normalisation and regularisation. This makes PQN an important comparison for Memory Merge DQN because the two methods approach the same stability problem from opposite directions. PQN simplifies value learning by removing the target network entirely, whereas Memory Merge keeps the target network but changes how it is updated. Comparing against PQN therefore tests whether improving the target update mechanism remains useful relative to a state of the art Q-learning baseline that avoids target network altogether.

Parameter averaging is not a new concept within deep learning, but has not been well-explored in deep reinforcement learning, the closest piece of work \cite{nikishin2018improving} uses Stochastic Weight Averaging (SWA) \citep{izmailov2018averaging} but was limited in its scope.  To the best of the knowledge of the authors, papers using parameter averaging to address stability in value-based learning remains limited.  More directly related to this paper is Fisher Weighted Model Merging introduced by \citet{matena2022merging}. In that work, Fisher Weighted Model Merging  is a technique in which parameters from multiple trained models are combined using importance weights derived from the Fisher information \citep{matena2022merging}. Intuitively, the hypothesis is  parameters that the model is more sensitive to are treated as more important and therefore receive greater influence in the merged model. This differs from simple uniform weight averaging, where every model contributes equally to every parameter. Memory Merge is related in spirit because it also performs a parameter weighted merge, but it differs in both the setting and importance estimator. Fisher Weighted Model Merging \cite{matena2022merging} is typically used to merge independently trained or fine tuned models using a Fisher estimate. In contrast, this paper's method performs repeated online merges of recent DQN copies specifically for target network updates, and its weights are based on the Q-value sensitivity on sample states from the replay buffer. Another piece of work worth discussing as it is similar in spirit is Memory Aware Synapses (MAS) \cite{aljundi2018memory} who used a similar method to identify parameter sensitivity between models in a deep learning setting, however unlike Memory Aware Synapses, which uses sensitivity estimates to protect important parameters from being changed during continual learning, Memory Merge uses Q-value sensitivity to weight how recent DQN parameter copies are merged into a single target network during bootstrapped value learning. Thus, Memory Merge method adapts the general idea of importance weighted parameter merging to the bootstrapped value learning setting.

\section{Memory Merge Update Rule}

\subsection{Memory Merge DQN}

\begin{figure*}
\centering
\includegraphics[scale=0.35]{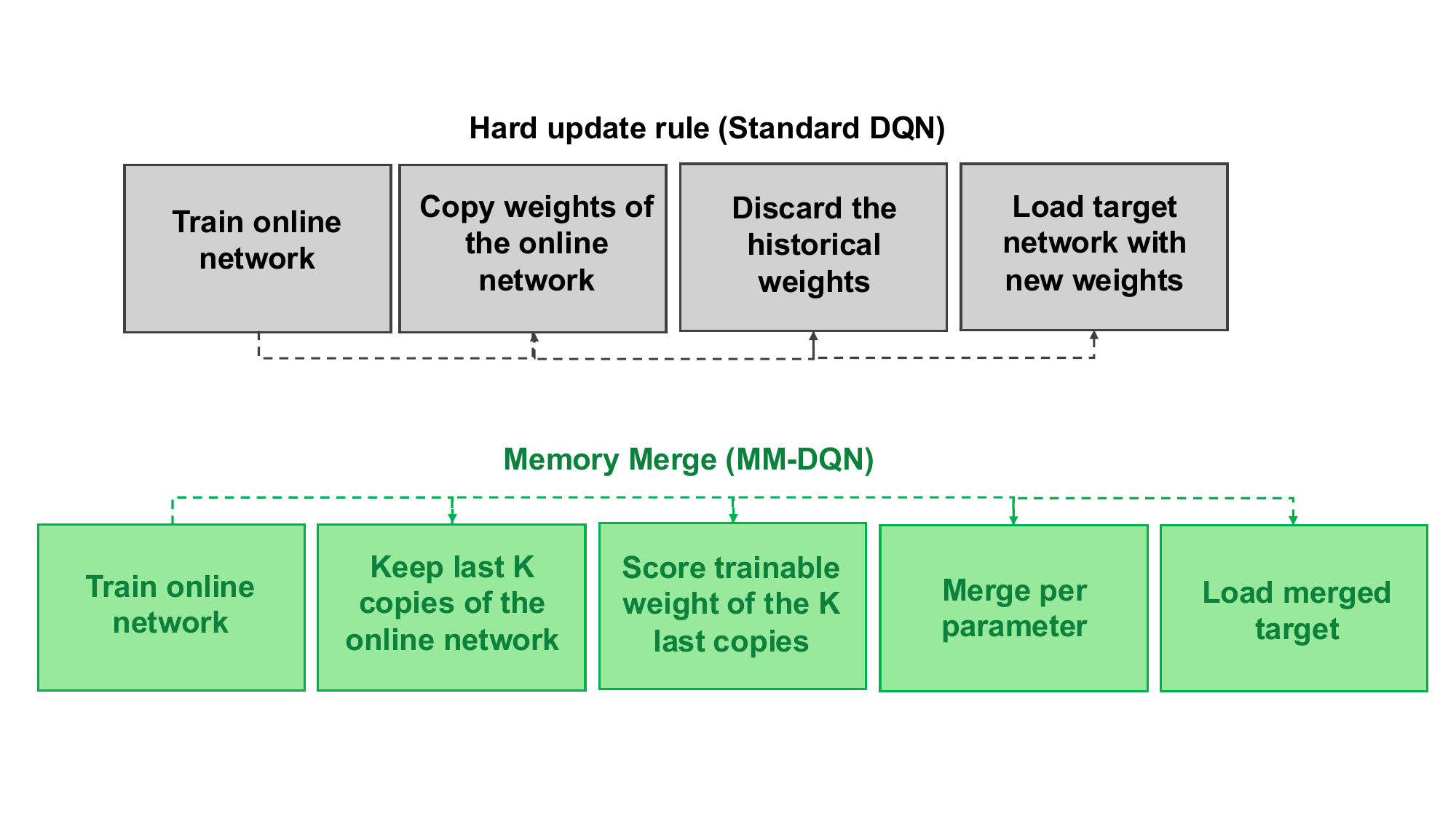}
\caption{Hard target update versus the Memory Merge update. Both train the online network for gradient steps between updates. Top refers to the DQN hard update copies the newest parameters into the target, discarding all intermediate history and shifting the bootstrap targets abruptly. The bottom shows the MM-DQN retains the K most recent copies, scores each parameter by Q sensitivity estimate on sample states, and loads their element wise weighted average with a recency prior toward the newest copy as the target. The merged target is a single network.}
\label{fig2:mode_fail}
\end{figure*}

This paper proposes a memory based target network update rule for DQN where the central idea is to give the target network access to a short history of recent online networks, rather than forcing it to depend only on the newest one. In standard DQN, the target network is periodically replaced by a direct copy of the latest online network, which may discard useful parameter information from earlier stages of learning. This proposed method instead maintains a small rolling memory of recent online network copies and uses this memory to construct the next target network. Parameters from historical copies are retained when they appear important for the current Q-value behaviour, producing a merged target network that combines recent learning with useful information from earlier online networks. The full algorithm is available in Algorithm~\ref{alg:memory_merge_update} and the difference between the update mechanisms are outlined in Figure \ref{fig2:mode_fail}.

The online Q-network is still trained using the standard DQN loss. For a transition $(s_t,a_t,r_t,s_{t+1},d_t)$ sampled from the replay buffer, the temporal difference target is
\begin{equation}
    y_t =
    r_t +
    \gamma (1-d_t)
    \max_{a'} Q_{\theta^-}(s_{t+1},a'),
\end{equation}
where $\theta$ denotes the online network parameters, $\theta^-$ denotes the target network parameters, $\gamma$ is the discount factor, and $d_t$ is the terminal signal. The factor $(1-d_t)$ removes the bootstrap term for terminal transitions.

The online network is trained by minimising the squared temporal difference error \cite{mnih2015human}
\begin{equation}
    L(\theta)
    =
    \mathbb{E}_{(s,a,r,s',d)\sim D}
    \left[
    \left(
    Q_{\theta}(s,a) - y
    \right)^2
    \right],
\end{equation}
where $D$ is the replay buffer. This proposed method does not change this loss and only changes how the target network parameters $\theta^-$ are updated.

At each scheduled target network update, the current online network parameters are saved into a finite rolling memory:
\begin{equation}
    M_t =
    \left(
    \theta^{(1)}, \theta^{(2)}, \ldots, \theta^{(K)}
    \right).
\end{equation}
The copies are ordered from oldest to newest. Therefore, $\theta^{(K)}$ is the newest online network copy, and $K \leq K_{\max}$, where $K_{\max}$ is the maximum number of stored copies.

If the memory contains only one copy, there is nothing to merge. In that case, the method uses the standard hard target update:
\begin{equation}
    \theta^- \leftarrow \theta^{(K)}.
\end{equation}

When at least two copies are available, the method builds the target network by merging the stored copies. The merge is guided by a small set of observations sampled from the replay buffer. Let
\begin{equation}
    P = \{x_i\}_{i=1}^{N}
\end{equation}
denote the retained temporal difference observations. Each $x_i$ is a next state observation retrieved from the replay buffer and terminal transitions are excluded before forming this sample set, this is because the target network only contributes to the bootstrap term for non-terminal transitions..

The newest copy chooses one anchor action for each sample observation:
\begin{equation}
    a_i^*
    =
    \arg\max_a Q_{\theta^{(K)}}(x_i,a).
\end{equation}
Here, $a_i^*$ is the greedy action selected by the newest copy for sample observation $x_i$. These same anchor actions are used to evaluate every stored copy. This means that older copies are evaluated using the actions preferred by the newest copy, rather than using their own greedy actions. This anchored action choice is important because it makes the sensitivity weights comparable across stored copies. If each copy were evaluated using its own greedy action, then the merge weights would reflect both parameter sensitivity and disagreement between older and newer policies, hence the need to fix the action.

The equation below is based on the Fisher Weighted Model Merging rule \cite{matena2022merging}, but it does not estimate a Fisher information matrix, instead, it uses a Q-value sensitivity measure. The purpose of this measure is to estimate how strongly each parameter affects the Q-values on the sample observations.

Here, $\ell$ refers to a particular parameter group in the network, such as the weights or biases of a layer, and $j$ refers to one individual number inside that parameter group. For copy $k$, the sensitivity of parameter element $\theta_{\ell,j}^{(k)}$ is
\begin{equation}
    W_{\ell,j}^{(k)}
    =
    \frac{1}{N}
    \sum_{i=1}^{N}
    \left(
    \frac{
    \partial Q_{\theta^{(k)}}(x_i,a_i^*)
    }{
    \partial \theta_{\ell,j}^{(k)}
    }
    \right)^2 .
\end{equation}
A larger value of $W_{\ell,j}^{(k)}$ means that this parameter element has a stronger local influence on the Q-values for copy $k$.

Thus, copy $k$ contributes more strongly to parameter element $j$ in parameter group $\ell$ when that parameter has higher Q-value sensitivity in that copy.

The merge also includes a small recency bias toward the newest copy. This bias is represented by $\lambda_\ell$, where $\lambda_\ell \geq 0$ controls how strongly parameter group $\ell$ is pulled toward the newest copy. The recency coefficient was not tuned as an additional performance component, it is fixed across all environments and used only as a small stabilising prior to prevent the sensitivity weighted merge from drifting too far from the newest online network. However, future work could focus on tuning this hyper-parameter.

The merged target network parameter is computed as
\begin{equation}
    \theta_{\ell,j}^{-}
    =
    \frac{
    \sum_{k=1}^{K}
    W_{\ell,j}^{(k)}
    \theta_{\ell,j}^{(k)}
    +
    \lambda_\ell
    \theta_{\ell,j}^{(K)}
    }{
    \sum_{k=1}^{K}
    W_{\ell,j}^{(k)}
    +
    \lambda_\ell
    }.
\end{equation}

This equation combines two sources of information. The first source is the memory of the different copies. Each stored copy contributes its parameter value $\theta_{\ell,j}^{(k)}$, and this contribution is scaled by its sensitivity weight $W_{\ell,j}^{(k)}$. A larger sensitivity weight means that the parameter value from that copy has more influence on the final target network parameter.

The second source is the newest online network copy, $\theta_{\ell,j}^{(K)}$. This copy already appears in the copy memory, but it is also given an extra weight $\lambda_\ell$. This extra weight acts as a recency bias, helping the merged target network stay closer to the most recent online network.

The denominator is the total amount of weight used in the merge. Dividing by this total weight keeps $\theta_{\ell,j}^{-}$ on the same scale as the stored parameter values. In other words, the update is a weighted average where older copies can still contribute when their parameters are important, but the newest copy is given an additional preference.

\subsection{Experimental Methods}

All experiments were implemented using the CleanRL framework \cite{huang2022cleanrl} and trained on the Atari learning environment through gymnasium \cite{towers2026gymnasium}. Each Atari environment was constructed with the standard value-based Atari preprocessing pipeline and were trained for 10 million environment steps. To reduce the effect of random seed selection, each algorithm was evaluated using four matched random seeds, with the same seed set used for every algorithm and every ablation variant. This ensures that differences between algorithms are not driven by one method receiving an easier set of stochastic initialisations.

All the algorithms with the exception of vanilla DQN follows the standard DQN convolutional structure commonly used for Atari value-based agents, but replaces the standard ReLU activation functions with GELU activation functions. Layer normalisation is applied after each convolutional layer and after the hidden linear layer, before the GELU activation function. The final linear layer outputs one Q-value for each discrete action. The full set of hyper-parameters can be found in Table~ \ref{tab:hyper}. Vanilla DQN uses ReLu activation function and no Layer normalisation. Additionally, PQN was the only algorithm to use gradient clipping, since this was included in the CleanRL implementation out of the box. This gave PQN a potential optimisation stability advantage relative to the other methods. However, we retained this default configuration because our goal was not only to compare against the published CleanRL PQN algorithm, but also to assess whether the proposed Memory Merge approach could improve upon a strong and practically tuned reference implementation.

Layer normalisation \cite{ba2016layer} is included for two related reasons. First, layer normalisation is particularly relevant for Memory Merge because the target network is constructed by merging parameters from recent online network copies. Parameter merging is meaningful only when the stored networks remain sufficiently compatible. If successive copies represent Q-values using very different hidden activation scales, then averaging or merging their parameters can produce a target network whose intermediate representations are poorly calibrated. Layer normalisation helps reduce this risk by normalising the activations within each layer, making the functional behaviour of nearby copies less dependent on arbitrary changes in feature magnitude. This does not guarantee perfect alignment across all stored networks, however, because Memory Merge combines temporally adjacent copies from the same training trajectory, layer normalisation provides a practical mechanism for keeping their representations more stable and comparable across target updates.

Second, normalisation has been shown to improve the stability of deep reinforcement learning and has been connected to reduced overestimation and more robust temporal difference learning \citep{nauman2024overestimation, palenicek2025scaling, bhatt2019crossnorm, gallici2024simplifying}. Consequently, layer normalisation acts as a stabiliser that makes the underlying DQN training procedure less sensitive to representation drift. This is important for the proposed method because Memory Merge depends on reusing previous parameter states rather than discarding them at every target update.

\begin{algorithm}[!htbp]
\caption{Memory Merge target network Update}
\label{alg:memory_merge_update}
\begin{algorithmic}[1]
\Require Online network parameters $\theta$, Target network parameters $\theta^-$, copy memory $M_t$ with maximum size $K_{\max}$, replay buffer $D$, number of sample observations $N$, recency-prior weight $\lambda_\ell > 0$ for each parameter group $\ell$

\State Add a copy of the current online network to memory, remove the oldest copy from $M_t$ if it is full

\If{$K < 2$}
    \State $\theta^- \leftarrow \theta^{(K)}$
    \State \Return
\EndIf

\State Sample non-terminal next-state observations from $D$:
\[
    P = \{x_i\}_{i=1}^{N}
\]

\State Compute anchor actions using the newest copy:
\[
    a_i^* \leftarrow \arg\max_a Q_{\theta^{(K)}}(x_i,a),
    \qquad x_i \in P
\]

\State Compute Q-value sensitivity weights:
\[
    W_{\ell,j}^{(k)}
    \leftarrow
    \frac{1}{N}
    \sum_{i=1}^{N}
    \left(
    \frac{
    \partial Q_{\theta^{(k)}}(x_i,a_i^*)
    }{
    \partial \theta_{\ell,j}^{(k)}
    }
    \right)^2
\]
for each copy $k$, parameter group $\ell$, and scalar parameter $j$

\State Update each target network parameter:
\[
    \theta_{\ell,j}^{-}
    \leftarrow
    \frac{
    \sum_{k=1}^{K}
    W_{\ell,j}^{(k)}
    \theta_{\ell,j}^{(k)}
    +
    \lambda_\ell
    \theta_{\ell,j}^{(K)}
    }{
    \sum_{k=1}^{K}
    W_{\ell,j}^{(k)}
    +
    \lambda_\ell
    }
\]
\end{algorithmic}
\end{algorithm}

\subsubsection{Memory size ablation}

To isolate the effect of the number of stored network copies, this paper conducted a memory size ablation study on two Atari games, Zaxxon and River Raid. In this experiment, all non-memory hyperparameters are held fixed and only the number of stored online network copies is varied. We compare six variants, 2, 3, 4, 5, 6 and 9 stored historical copies. Each variant is evaluated using the same four matched random seeds, and the plotted shaded region corresponds to one standard deviation across seeds.

\begin{figure}[t]
\centering
\includegraphics[width=\textwidth]{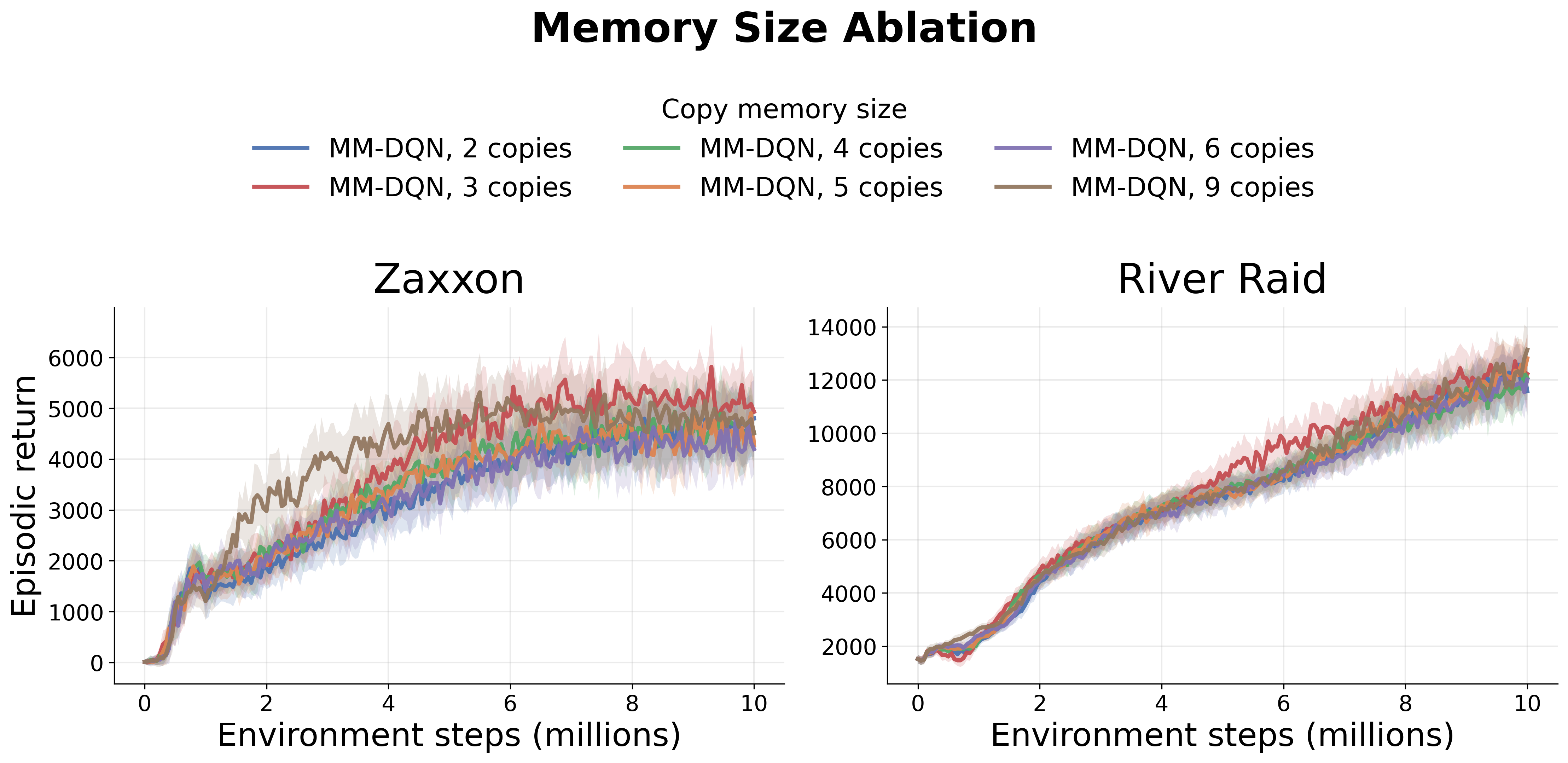}
\caption{Memory size ablation for Memory Merge DQN on Zaxxon and River Raid. Each curve shows the mean episodic return over four matched seeds, and the shaded region shows one standard deviation. The variants differ only in the number of stored online network copies used to construct the merged target network. A moderate memory size of $K=3$ is consistently competitive, while larger memories can accelerate early learning but do not always improve final performance.}
\label{fig:memory_merge_copy_ablation}
\end{figure}

Figure~\ref{fig:memory_merge_copy_ablation} shows that the copy memory size affects both learning speed and final performance, but that the effect is game dependent. In Zaxxon, the largest memory setting, $K=9$, learns fastest during the early and middle stages of training. This suggests that a longer memory can help the target network reuse earlier parameter information and accelerate the formation of a useful value function. However, this early advantage does not translate into a clearly superior final policy. By the end of training, the $K=3$ setting reaches the highest or near highest return, while the $K=4$, $K=5$, and $K=9$ variants form a similar upper group. The smaller $K=2$ setting and the $K=6$ setting are generally lower. This indicates that increasing the copy memory can improve early learning, but that too much retained history may eventually become less useful if older parameters become less compatible with the current value function.

In River Raid, the differences between memory sizes are smaller. All variants improve steadily throughout training and the standard deviation bands overlap for much of the run. Nevertheless, the $K=3$ setting is consistently among the strongest variants during the middle and late stages of training, with $K=9$ also competitive near the end. This suggests that River Raid is less sensitive to the exact memory size than Zaxxon, provided that the method has access to more than the minimal two copy. The result is important because it shows that the proposed method does not require a very large memory to work well.

Overall, the ablation supports the choice of $K=3$ as the default Memory Merge configuration. A two copy memory can be too short to capture enough useful historical information, while very large memories can increase computational cost and may introduce older parameter states that are less aligned with the current policy. The best setting is therefore not necessarily the largest memory. Instead, the results suggest that Memory Merge benefits from a short but informative history.

\section{Results}
\subsection{Final performance}

To evaluate final performance, we compute the average episodic return over the final 200K environment steps for each game and algorithm. This evaluation window measures performance near the end of training. We compare Memory Merge DQN, denoted as MM-DQN, against vanilla DQN, Averaged DQN, DQN with layer normalisation, and PQN (with gradient clipping) across 54 Atari environments. The raw scores used for this analysis are reported in Table~\ref{tab:appendix_final_200k_scores}.

Because Atari scores vary substantially in scale across games, the raw average score across environments is not used as the primary comparison metric. Instead, we report game level rankings, pairwise wins, and large improvement counts relative to vanilla DQN. These metrics are intended to measure whether an algorithm improves performance consistently across games, rather than allowing a small number of large score environments to dominate the comparison.

Table~\ref{tab:final_200k_rank_summary} summarises the aggregate ranking results across the Atari game suite. Across the 54 evaluated games, MM-DQN obtains the largest number of first place results, with 21 wins. PQN (with gradient clipping) is close behind with 20 wins, followed by Averaged DQN with 7 wins, DQN with layer normalisation with 6 wins, and vanilla DQN with 0 wins. This indicates that Memory Merge is highly competitive as a final performance method and obtains the strongest first place count among the evaluated methods.

\begin{table}[t]
\centering
\caption{Aggregate ranking performance over the final 200K environment steps. Lower mean rank is better. Percentages are computed over 54 games. Ties are handled with average ranks and last place counts include all methods tied for the lowest score on a game.}
\label{tab:final_200k_rank_summary}
\begin{tabular}{lrrrrrr}
\toprule
Algorithm & Wins & Win rate & Top 2 & Top 3 & Last place & Mean rank \\
\midrule
Averaged DQN & 7 & 13.0\% & 16 & 31 & 7 & 3.11 \\
DQN with layer norm & 6 & 11.1\% & 26 & \textbf{43} & \textbf{1} & 2.63 \\
MM-DQN & \textbf{21} & \textbf{38.9\%} & 28 & 34 & 10 & 2.65 \\
PQN & 20 & 37.0\% & \textbf{34} & 38 & 14 & \textbf{2.54} \\
DQN & 0 & 0.0\% & 4 & 15 & 24 & 4.07 \\
\bottomrule
\end{tabular}
\end{table}

These results should also be interpreted in light of an important implementation difference. PQN was the only method in our comparison that used gradient clipping, inherited from the default CleanRL implementation. Since gradient clipping can improve optimisation stability by limiting large updates, PQN had an additional stabilising mechanism that was not present in MM-DQN, Averaged DQN, DQN with layer normalisation, or vanilla DQN. We therefore treat PQN as a particularly strong out-of-the-box baseline rather than as a purely architecture matched comparison. Under this setting, the fact that MM-DQN obtains the largest number of first place results, with 21 wins compared with 20 for PQN (with gradient clipping), is notable. Memory Merge remains highly competitive despite not using the extra stabilisation provided by gradient clipping.

Table~\ref{tab:mmdqn_pairwise} provides the pairwise comparison between MM-DQN and each baseline. MM-DQN outperforms vanilla DQN on 41 out of 54 games. It also records more wins than losses against Averaged DQN and DQN with layer normalisation, with 29 wins and 25 losses in both comparisons. Against PQN (with gradient clipping), MM-DQN is slightly ahead in pairwise count, with 28 wins and 26 losses.  At the same time, PQN remains highly competitive under the aggregate ranking view in Table~\ref{tab:final_200k_rank_summary}, where it achieves the best mean rank and the most Top 2 finishes.

\begin{figure*}[!htpb]
    \centering
    \includegraphics[width=\textwidth]{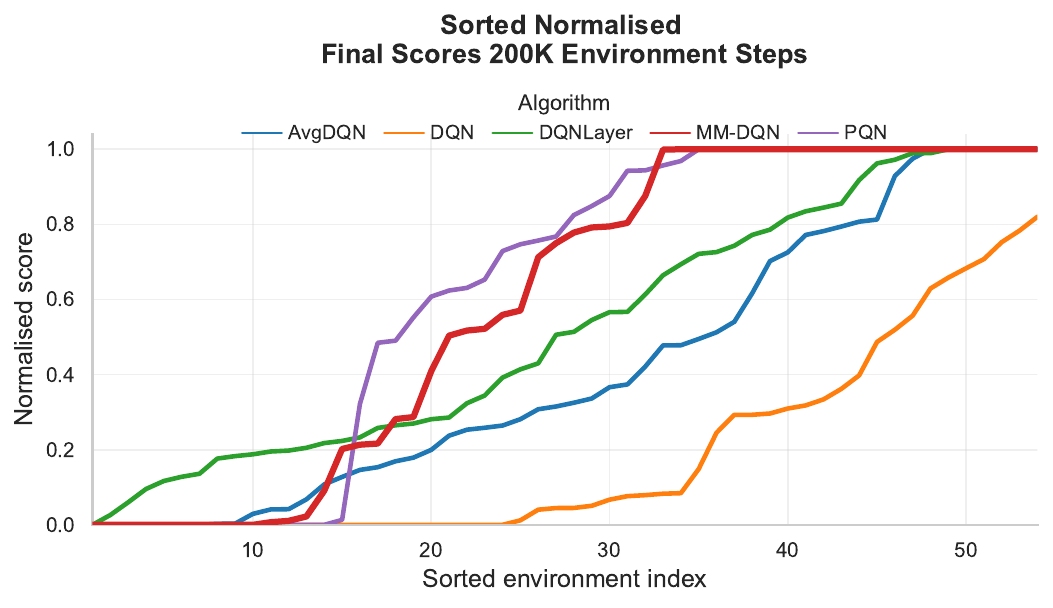}
    \caption{
        Sorted normalised final performance scores across the 54 Atari environments.
        For each environment, algorithm scores are normalised relative to the worst and best score obtained by any evaluated method on that environment.
        Each algorithm's 54 normalised scores are then sorted independently from lowest to highest before plotting.
        The x-axis is therefore a sorted environment index rather than a fixed game order, so curves should be interpreted as distributions of relative performance across environments.
        Higher curves indicate that an algorithm remains closer to the best observed method over a larger fraction of environments.}
    \label{fig:normalised_sorted_scores}
\end{figure*}

As an additional scale normalised view, Figure~\ref{fig:normalised_sorted_scores} plots the empirical distribution of within game normalised final scores. For each environment, the scores of all algorithms are normalised relative to the worst and best score achieved on that game, so that 0 corresponds to the weakest method and 1 corresponds to the strongest method for that environment. The normalised scores for each algorithm are then sorted independently before plotting. This visualisation does not preserve game identity and should therefore not be interpreted as a game by game trajectory. Instead, it shows how often each method remains close to the best method after removing the effect of Atari score scale. Under this view, MM-DQN forms one of the strongest curves, remaining close to the top of the normalised range over a large fraction of games, while PQN (with gradient clipping) is also highly competitive and often slightly more consistent across the distribution. Vanilla DQN is clearly lower for most of the curve, supporting the conclusion that the proposed target update improves final performance beyond the standard hard copy baseline.

\begin{table}[t]
\centering
\caption{Pairwise game counts for MM-DQN against each baseline over the final 200K environment steps.}
\label{tab:mmdqn_pairwise}
\begin{tabular}{lrrr}
\toprule
Comparison & MM-DQN better & MM-DQN worse & MM-DQN win rate \\
\midrule
MM-DQN vs DQN & \textbf{41} & 13 & 75.9\% \\
MM-DQN vs Averaged DQN & \textbf{29} & 25 & 53.7\% \\
MM-DQN vs DQN + LayerNorm & \textbf{29} & 25 & 53.7\% \\
MM-DQN vs PQN & \textbf{28} & 26 & 51.9\% \\
\bottomrule
\end{tabular}
\end{table}

Several of the strongest MM-DQN wins are substantial when compared with the next best method in the same game. For example (refer to Table~\ref{tab:appendix_final_200k_scores}), MM-DQN improves over the next best method by approximately 66.7\% on Asterix, 39.5\% on Beam Rider, 45.4\% on Defender, 45.8\% on Phoenix, and 18.0\% on Zaxxon. These cases indicate that Memory Merge can do more than smooth training and in some environments, it substantially changes the final policy reached.

\begin{table}[t]
\centering
\caption{Large improvements over vanilla DQN. A game is counted if the algorithm improves over DQN by at least 50\%. Percentages are computed over 54 games.}
\label{tab:large_improvements_vs_dqn}
\begin{tabular}{lrrr}
\toprule
Algorithm & Games $\geq 50\%$ better than DQN & Percentage & Also best overall \\
\midrule
Averaged DQN & 9 & 16.7\% & 2 \\
DQN + LayerNorm & 15 & 27.8\% & 2 \\
MM-DQN & 20 & 37.0\% & \textbf{15} \\
PQN & \textbf{21} & \textbf{38.9\%} & 13 \\
\bottomrule
\end{tabular}
\end{table}

We also examine large improvements relative to vanilla DQN. For each algorithm, we count the number of games where the final score is at least 50\% better than DQN. Table~\ref{tab:large_improvements_vs_dqn} shows that PQN (with gradient clipping) has one more large improvement over DQN than MM-DQN. However, MM-DQN converts these large improvements into first place results more often. Of the 20 games where MM-DQN is at least 50\% better than DQN, it is also the best performing method overall on 15 games. PQN (with gradient clipping) is best overall on 13 of its 21 large improvement games, while Averaged DQN and DQN with layer normalisation are best overall on only 2 such games each. This suggests that when Memory Merge improves over the standard DQN baseline, the improvement is often large enough to exceed the other baselines as well.

There are six games where MM-DQN is the only method that improves over DQN by at least 50\% under the definition above (Alien, Asterix, Beam Rider, Phoenix, River Raid, and Skiing). These cases are especially important because they suggest that Memory Merge is not merely benefiting from general architectural changes shared by the other baselines. Instead, the sensitivity weighted merge can produce distinct gains that are not matched by uniform averaging, layer normalisation alone, or state of the art algorithms such as PQN (with gradient clipping).

The results are also informative in environments with sparse rewards or delayed credit assignment. MM-DQN also performs strongly on Elevator Action, increasing from 165.62 for DQN to 4029.55, and on Skiing, where it improves from $-28757.28$ to $-13264.81$. These results are consistent with the motivation that preserving useful value function history can help stabilise learning in environments where reward information is sparse, delayed, or difficult to propagate. The value scale behaviour associated with these final scores is examined further in Figure~\ref{fig:q}.

Overall, the results support the main hypothesis of this paper, where replacing hard target copying with a Q-value sensitivity weighted merge over recent online network copies can improve the final performance of DQN agents, and while it is not uniformly the best method across every aggregate metric, it is competitive with state of the art value-based algorithms like PQN (with gradient clipping) and provides a competitive alternative in value-based learning.

\subsection{Value scale alignment after Memory Merge}

The merged weights within MM-DQN are shaped by the sensitivity of parameters to selected Q-values, hence, Memory Merge should affect not only final episodic return but also the scale and reliability of the value estimates used for bootstrapping. A useful merge should preserve high value structure when it supports strong behaviour, while avoiding target updates that retain large but poorly supported Q-values. 

The final score results show that Memory Merge often improves policy quality, but these do not by explain how the target update changes value learning. Since the proposed method constructs the target network using Q-value sensitivity, an important question is whether the resulting value estimates remain behaviourally meaningful. We therefore examine whether late training predicted Q-values are aligned with realised episodic returns. This analysis connects the performance gains above with the central mechanism of the method,  preserving useful value-function structure without retaining misleading target network information.

\begin{figure}[t]
\centering
\includegraphics[width=\textwidth]{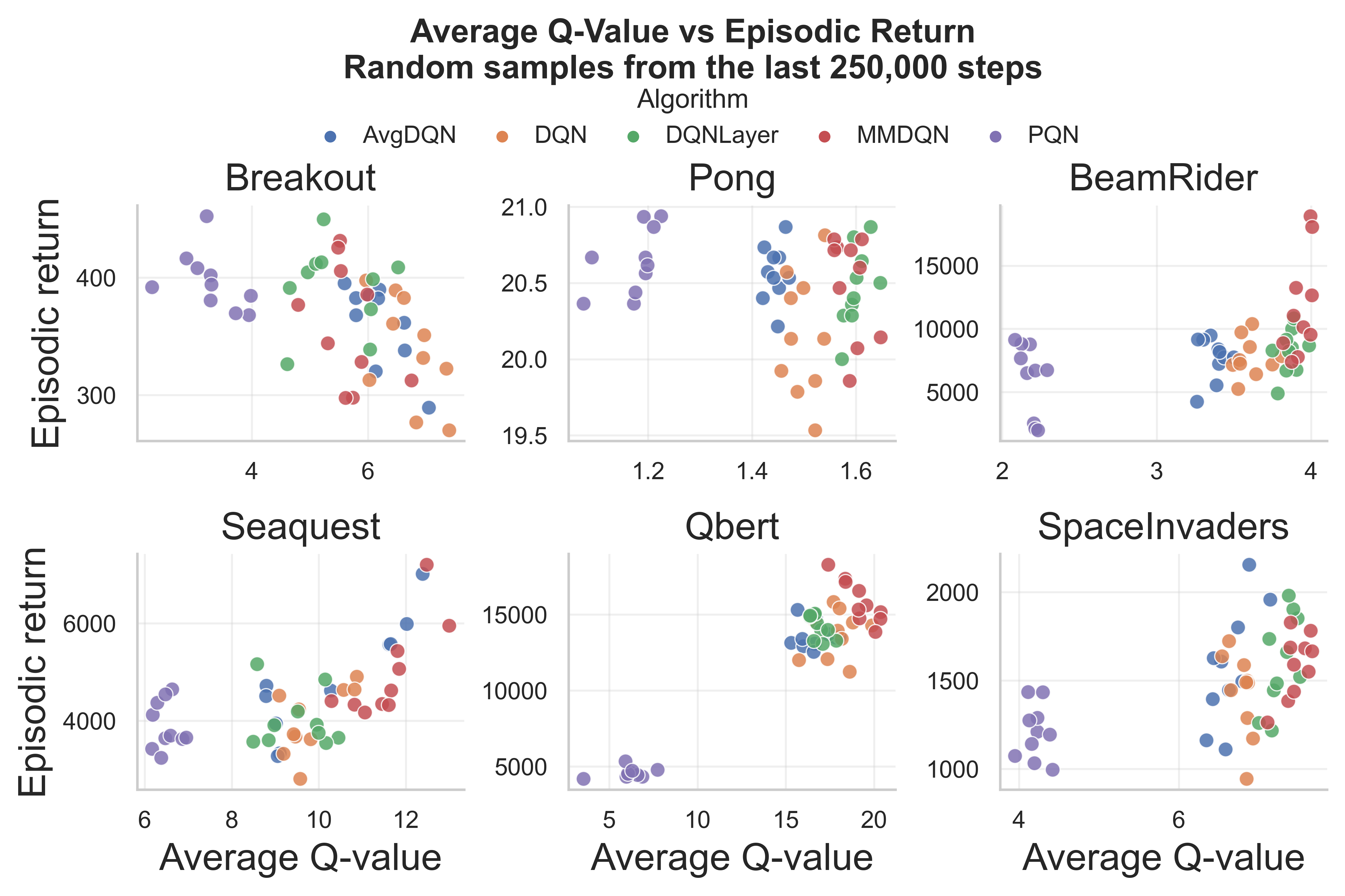}
\caption{Average predicted Q-value versus realised episodic return for random samples from the final 250K environment steps. Each panel shows one Atari game and each colour corresponds to one algorithm. Points with high average Q-values but low episodic returns indicate value miscalibration.}
\label{fig:q}
\end{figure}

The preceding results show that Memory Merge can improve final performance, but the purpose of the method is not only to increase return. Its aim is to construct a target network that preserves useful value-function structure while limiting the influence of outdated or misleading parameter weights. Because DQN learns through bootstrapped targets, errors in Q-value  can be repeatedly copied into later updates. The proposed merge rule is therefore expected to affect the relationship between predicted Q-values and realised episodic returns. To examine this relationship, Figure~\ref{fig:q} compares average predicted Q-values with realised episodic returns during late training for Breakout, Pong, BeamRider, Seaquest, Qbert, and SpaceInvaders. These games are drawn from the original DQN environment set \cite{mnih2013playing}, making this diagnostic directly relevant to environments historically used to evaluate DQN agents. 

This diagnostic is especially relevant to Memory Merge because the method does not simply average returns or smooth predictions. It merges parameters according to how strongly they influence current Q-values. If the sensitivity weights select useful historical structure, then MM-DQN should be able to maintain large value estimates in states where those estimates are supported by strong subsequent performance. Conversely, if the merge retains obsolete parameter, then high predicted Q-values may appear without corresponding episodic returns. Figure~\ref{fig:q} should therefore be interpreted as a check on when target network memory produces meaningful value estimates and when value scale remains environment dependent.

The relationship between Q-value scale and performance is strongly environment dependent. In Breakout, the association is weak and non-monotonic. PQN (with gradient clipping) obtains relatively high returns at lower Q-values, while DQN, Averaged DQN, DQN with layer normalisation, and MM-DQN occupy overlapping higher Q regions with a broad range of returns. Several samples with larger Q-values do not correspond to higher episodic returns, indicating that value magnitude alone is not a reliable indicator of policy quality in this game. Pong also provides limited calibration information because returns are tightly concentrated near the maximum score. Most algorithms achieve returns around 20--21 despite noticeably different Q-value ranges, suggesting a ceiling effect where episodic performance saturates and differences in Q-value scale become less informative.

BeamRider, Seaquest, and Qbert show a more positive association between Q-values and episodic return. In BeamRider, PQN (with gradient clipping) occupies a lower Q-value range and generally lower return region, while MM-DQN reaches the highest observed returns among the plotted samples and does so at comparatively high Q-values. Seaquest shows a similar trend where MM-DQN is concentrated toward the higher Q-values and higher return region, although there remains substantial overlap and variance among algorithms. Qbert exhibits the clearest separation, with PQN (with gradient clipping) clustered at low Q-values and low returns, while the DQN-based methods occupy much higher Q-value and return regions. Within this high-performing region, MM-DQN frequently appears near the upper end of the return distribution.

The BeamRider and Qbert plots are particularly important because they connect the final performance results with the proposed mechanism. In these environments, MM-DQN reaches high return regions while also maintaining high predicted Q-values, suggesting that the method is not merely damping value estimates but preserving value scale when that scale remains useful for behaviour. SpaceInvaders is more mixed where higher Q-values generally correspond to higher returns. MM-DQN lies in a high Q-values region with competitive returns, but Averaged DQN and DQN with layer normalisation produced a mixture of high Q-values and lower returns with high Q-values and high returns. 

Overall, Figure~\ref{fig:q} supports the interpretation that Memory Merge can alter the value-learning dynamics of the DQN agent, and does not simply suppress Q-values. In several environments, particularly BeamRider, Seaquest, and Qbert, MM-DQN was able to maintain high value estimates as well as strong realised returns. This is consistent with the intended role of the Q-value sensitivity weighted merge which is to retain parameter information that remains important for current Q-value behaviour. 

\section{Conclusion}
Memory Merge DQN replaces hard target copying with a  Q-value sensitivity weighted merge over recent online network copies. Across 54 Atari games, the method achieves the largest number of first place final scores, outperforms vanilla DQN, and is highly competitive with state of the art modern DQN algorithms like PQN (with gradient clipping). These results suggest that selectively constructing the target network can be useful, but only when older parameter weights remain compatible with the current value function. Future work can expand on this work by focusing on applying this across into actor-critic methods or combining it with other DQN extensions.

\noindent

\bibliography{sn-bibliography}

\appendix
\section{Appendix}
\begin{longtable}{@{}llllll@{}}
\caption{Final 200K time-steps Atari Scores}
\label{tab:appendix_final_200k_scores}\\
\toprule
Games & \pbox{20cm}{Averaged \\DQN} & DQN & \pbox{20cm}{DQN \\(LayerNorm)} & MM-DQN & PQN \\* 
\midrule
\endfirsthead

\toprule
Games & \pbox{20cm}{Averaged \\DQN} & DQN & \pbox{20cm}{DQN \\(LayerNorm)} & MM-DQN & PQN \\* 
\midrule
\endhead
\bottomrule
\endfoot
\endlastfoot
Alien & 1333.96 & 1304.18 & 1157.9 & 2141.83 & 1634.87 \\
Amidar & 464.29 & 283.2 & 462.41 & 412.22 & 436.9 \\
Assault & 3099.89 & 3607.58 & 2806.64 & 2179.64 & 6789.94 \\
Asterix & 16832.23 & 11387.87 & 6742.28 & 28059.08 & 3615.62 \\
Asteroids & 1023.72 & 985.48 & 1016.79 & 986.34 & 1106.83 \\
BankHeist & 774.69 & 462.89 & 792.8 & 711.1 & 906.83 \\
BattleZone & 18655.6 & 21543.09 & 24603.75 & 8030.63 & 29499.99 \\
BeamRider & 6899.13 & 7174.12 & 8516.91 & 11883.65 & 6748.58 \\
Berzerk & 603.16 & 564.06 & 611.46 & 795.08 & 782.09 \\
Bowling & 25.39 & 25.29 & 42.03 & 55.99 & 47.67 \\
Boxing & 88.96 & 87.62 & 92.57 & 96.35 & 95.85 \\
Breakout & 359.49 & 344.32 & 401.7 & 355.91 & 399.23 \\
Carnival & 5049.12 & 4160.58 & 4285.14 & 4790.91 & 3792.31 \\
Centipede & 3708.53 & 3236.08 & 3971.64 & 3368.18 & 4689.28 \\
ChopperCommand & 2218.82 & 2204.64 & 6669.79 & 7553.13 & 4827.51 \\
CrazyClimber & 107998.78 & 99941.21 & 117592.37 & 112693.17 & 131114.06 \\
Defender & 3369.61 & 3527.39 & 4898.24 & 16467.45 & 11329.35 \\
DemonAttack & 26716.24 & 30062.57 & 39771.26 & 40803.36 & 92735.82 \\
DoubleDunk & -10.81 & -12.78 & -16.63 & -14.62 & -18.78 \\
ElevatorAction & 3.12 & 165.62 & 739.77 & 4029.55 & 56.25 \\
FishingDerby & -2.73 & -0.68 & 2.14 & 4.71 & 23.2 \\
Freeway & 24.92 & 22.07 & 33.32 & 33.31 & 31.92 \\
Frostbite & 3625.67 & 3322.38 & 4406.27 & 3647.22 & 985 \\
Gopher & 18279.38 & 11472.72 & 13850.15 & 10659.26 & 28748.13 \\
Hero & 4126.69 & 2306.21 & 13526.63 & 16582.24 & 11313 \\
IceHockey & -12.83 & -12.14 & -5.35 & -12.6 & -2.53 \\
Jamesbond & 588.47 & 590.03 & 676.13 & 677.9 & 506.64 \\
JourneyEscape & -12179.85 & -3210.62 & -1671.61 & -2158.06 & -733.88 \\
Kangaroo & 8612.56 & 9957.58 & 8494.55 & 6622.29 & 15002.88 \\
Krull & 8861 & 8556.09 & 8568.9 & 7209.05 & 9255.6 \\
KungFuMaster & 17760.8 & 376.84 & 24490.43 & 14580.56 & 28570.98 \\
MontezumaRevenge & 0 & 0 & 2.08 & 34.72 & 0 \\
MsPacman & 2107.17 & 2092.05 & 2281.43 & 2302.31 & 2460.56 \\
NameThisGame & 6698.13 & 6405.98 & 6217.23 & 4430.52 & 11172.15 \\
Phoenix & 8249.02 & 7733.98 & 8516.65 & 12414.79 & 5761.65 \\
Pitfall & -374.42 & -191.62 & -85.31 & -418.26 & -11.32 \\
Pong & 20.57 & 20.31 & 20.39 & 20.52 & 20.59 \\
Qbert & 13199.56 & 12989.72 & 14598.02 & 16422.21 & 4679.63 \\
Riverraid & 8706.56 & 8074.89 & 11071.28 & 12393.71 & 11301.03 \\
RoadRunner & 39002.39 & 35112.84 & 36206.84 & 37124.85 & 37539.36 \\
Robotank & 36.67 & 29.84 & 33.17 & 27.62 & 33.53 \\
Seaquest & 4713.76 & 3880.76 & 3897.9 & 4536.86 & 3811.88 \\
Skiing & -29035.89 & -28757.28 & -26244.49 & -13264.81 & -30184.04 \\
Solaris & 1292.37 & 783.61 & 1059.07 & 1424.17 & 989.88 \\
SpaceInvaders & 1549.82 & 1370.19 & 1559.64 & 1477.56 & 1165.41 \\
StarGunner & 41851.31 & 36271.49 & 32638.72 & 25840.8 & 57067.1 \\
Tennis & -13.03 & -13.59 & -3.9 & -0.16 & -3.28 \\
TimePilot & 4758.7 & 4191.84 & 4222.85 & 5377.33 & 4846.48 \\
Tutankham & 201.35 & 169.95 & 208.58 & 180.83 & 207.37 \\
Venture & 461.1 & 168.32 & 150.74 & 7.03 & 2.17 \\
VideoPinball & 230367.73 & 156513.73 & 369383.16 & 146249.65 & 375789.75 \\
WizardOfWor & 2320 & 913.99 & 5912.48 & 1995.7 & 5037.74 \\
YarsRevenge & 20306.16 & 20751.66 & 22399.47 & 22836.22 & 11227.91 \\
Zaxxon & 2911.45 & 1918.63 & 2548.54 & 5144.35 & 4360.53 \\* \bottomrule
\end{longtable}

\begin{longtable}[c]{@{}llllll@{}}
\caption{Hyper-parameters used. All value-based agents use the standard Atari convolutional architecture with three convolutional layers, $32$ filters with an $8 \times 8$ kernel and stride $4$, $64$ filters with a $4 \times 4$ kernel and stride $2$, and $64$ filters with a $3 \times 3$ kernel and stride $1$, followed by a $512$ unit fully connected layer and a Q-value output layer.}
\label{tab:hyper}\\
\toprule
Hyper-parameters            & DQN           & DQN LayerNorm & Averaged DQN & MM-DQN      & PQN                \\* \midrule
\endfirsthead
\endhead
\bottomrule
\endfoot
\endlastfoot
Timesteps                   & 10M   & 10M    & 10M   & 10M & 10M         \\
Gradient clipping      & No            & No            & No           & No         & Maxnorm 10.0 \\
Learning rate               & 0.0001        & 0.0001        & 0.0001       & 0.0001     & 0.0001             \\
Rollout/batch size &
  32 &
 32 &
   32 &
   32 &
   1,024 \\
Replay size          & 1M    & 1M     & 1M    & 1M  & None   \\
Gamma                       & 0.99          & 0.99          & 0.99         & 0.99       & 0.99               \\
Optimiser                   & Adam          & Adam          & Adam         & Adam       & Adam               \\
Activation & ReLU & GELU & GELU & GELU & GELU \\
Memory size     &           &           &          &     3   &         \\
Prior lambda     &           &           &          &     0.1   &         \\
Sensitivity sample     &           &           &          &     32   &         \\
Target update     & 1K          & 1K          & 1K         & 1K       & No target\\
No. Networks          &               &               & 2            &            &                       \\* \bottomrule
\end{longtable}

\end{document}